\documentclass[12pt]{article}
\usepackage[margin=1in]{geometry}
\usepackage{graphicx}
\usepackage{amsmath}
\usepackage{amssymb}
\usepackage{array}
\usepackage{hyperref}
\usepackage{pdflscape}
\usepackage{booktabs}
\usepackage{graphicx}
\usepackage{caption}
\usepackage{float}
\usepackage{hyperref}
\usepackage{authblk}
\title{Scaling Technology Acceptance Analysis with Large Language Model (LLM) Annotation Systems}
\author[1]{Pawel Robert Smolinski\thanks{Corresponding author: \texttt{pawel.smolinski@phdstud.ug.edu.pl}}}
\author[2]{Joseph Januszewicz}
\author[1]{Jacek Winiarski}
\affil[1]{\small University of Gdansk, Faculty of Economics, Sopot, Poland}
\affil[2]{\small Geisel School of Medicine at Dartmouth, Hanover, New Hampshire, United States}
\date{}
\begin{document}
\maketitle
\begin{abstract}
Technology acceptance models effectively predict how users will adopt new technology products. Traditional surveys, often expensive and cumbersome, are commonly used for this assessment. As an alternative to surveys, we explore the use of large language models for annotating online user-generated content, like digital reviews and comments. Our research involved designing an LLM annotation system that transforms reviews into structured data based on the Unified Theory of Acceptance and Use of Technology model. We conducted two studies to validate the consistency and accuracy of the annotations. Results showed moderate-to-strong consistency of LLM annotation systems, improving further by lowering the model temperature. LLM annotations achieved close agreement with human expert annotations and outperformed the agreement between experts for UTAUT variables. These results suggest that LLMs can be an effective tool for analyzing user sentiment, offering a practical alternative to traditional survey methods and enabling deeper insights into technology design and adoption.
\end{abstract}

\vspace{1cm}
\section{Introduction}
\footnotetext[0]{This is a preprint of a paper accepted for the 32nd International Conference on Information Systems Development (ISD 2024), 26--28/08/2024, Gdańsk, Poland}
Technology acceptance is a recognized concept within the field of information science \cite{davis1989perceived, venkatesh2003user}. It refers to the degree to which individuals perceive a new technology as useful and easy to use, and decide to use it \cite{davis1989perceived}. Understanding technology acceptance is crucial for developers, businesses \cite{thong1999integrated}, and policymakers \cite{kim2014international,smolinski2023nuclear} to predict how new technologies will be adopted by target users. To translate this understanding into actionable insights, researchers have developed technology acceptance models. Technology acceptance models facilitate the collection of data for refining technology design, assessing and improving organizational readiness for new IT solutions, and conducting SWOT (Strengths, Weaknesses, Opportunities, Threats) analysis of emerging technology products in market research.

Technology acceptance data is typically collected through surveys and questionnaires that assess respondents' perceptions of the usefulness, ease of use, social influence, and facilitating conditions related to the technology. These instruments are often based on the constructs defined in the TAM model and its variations (e.g. UTAUT). Responses are analyzed to predict attitudes toward the technology, intention to use it, and sometimes actual usage behavior. The results of such surveys provide valuable data for designing \cite{diamond2018using}, implementing \cite{park2015consumer}, and marketing \cite{robinson2005empirical} new technologies to improve their acceptance and adoption.

However, designing questionnaires to measure technology acceptance can present many challenges, often requiring substantial effort and resources. A primary obstacle is the lack of standardization across technology acceptance questionnaires \cite{chin2008fast,smolinski2023determinants}. Due to questionnaire heterogeneity, items researchers choose to represent key acceptance variables, such as Effort Expectancy, can differ widely from one study to another \cite{lewis2019comparison}. Moreover, collecting high-quality data on users' attitudes through questionnaires is a cumbersome process. It involves defining the target demographic for a given technological product, successfully reaching this group, and obtaining a sufficiently large sample size to ensure the reliability and validity of the TAM model. These steps are crucial for the accurate assessment of technology acceptance but can be challenging to execute in practice.

The questionnaires themselves also possess general limitations that restrict their utility in technology acceptance research. For example, respondents may provide socially desirable answers rather than truthful responses. Furthermore, participants might interpret the questions differently and questionnaire items often fail to capture the complexity of human attitudes, leading to inconsistent data \cite{chan2008so}.

These challenges in questionnaire design, data collection, and accurate attitude inferences have prompted a move within technology acceptance research towards leveraging more natural or ecologically valid data sources for inferring attitudes \cite{barki2007quo}. Ecologically valid data refers to information gathered from real-world settings or contexts that resemble the actual environment in which the phenomenon of interest naturally occurs. In the case of technology acceptance research, ecologically valid data sources include user-generated content such as social media posts, product reviews, and forum discussions, which reflect users' genuine attitudes and experiences with technologies in their everyday lives. The analysis of ecologically valid data has been accelerated by the development of Natural Language Processing (NLP) methods. NLP methods, such as sentiment analysis and Latent Dirichlet Allocation \cite{blei2003latent}, enable computers to understand, interpret, and generate human language in meaningful ways. By applying NLP, researchers can analyze vast amounts of unstructured textual content---such as social media posts, product reviews, and forum discussions---to extract data about users' attitudes toward technologies without several limitations of traditional survey methods \cite{grimmer2013text}. For example, Liu Zhuchenyang (2022) was the first to use NLP to extract data from app store reviews in order to validate a Technology Acceptance Model (TAM) \cite{liu2022user}. Their work demonstrated the potential of incorporating NLP-driven data into information systems and technology acceptance research.

In this paper, we aim to address the challenges associated with survey-based data collection in information system research by exploring the potential of Large Language Model (LLM) annotation systems as a consistent and scalable means of capturing user attitudes towards technology. Our research objectives include designing an LLM annotation system that converts unstructured user reviews into structured annotations and validating its consistency and accuracy. Through these objectives, we aim to demonstrate the potential of LLM annotation systems in replacing traditional questionnaire-based surveys and expert evaluations in technology and information science research.
\section{Large Language Model (LLM) annotation systems}
LLM annotation systems represent an emerging innovation in the field of Natural Language Processing \cite{kuzman2023chatgpt,kim2024meganno}. These systems leverage the capabilities of Large Language Models to convert textual data into a numerical format that can then be used for further statistical analysis \cite{liu2022user,kangale2016mining}. For instance, researchers interested in understanding how customer attitudes impact product market performance might traditionally rely on expensive market surveys, expert consultations, or focus groups. Now with increased proliferation of online data, researchers have the option to extract attitudes from customer reviews found on various forums and websites like Amazon. However, these textual reviews are initially in a text format and must be converted into a numerical format to be analyzable in statistical models, such as regression models. NLP annotation mechanisms serve this purpose by translating text into sentiment scores in the case of sentiment analysis and into any desired numerical output in the case of LLM annotation systems.

Although LLM annotation systems are relatively new and thus not extensively tested, there is a growing body of research exploring their potential \cite{tornberg2023chatgpt,reiss2023testing}. Several studies have already shown the promise of GPT models in classifying unseen data without prior training on specific tasks, such as detecting hate speech \cite{huang2023chatgpt}, identifying misinformation \cite{hoes2023leveraging}, assessing the credibility of news sources \cite{wei2024long}, and medical reports \cite{goel2023llms}. A handful of researchers have also begun to validate the effectiveness of LLM annotations in accurately inferring attitudes from textual data, showing promising results \cite{kuzman2023chatgpt,wei2024long}. Despite these initial successes, the overarching consensus is that further research is needed to fully understand and optimize the use of LLM annotation systems in attitude analysis and beyond \cite{tornberg2024best,reiss2023testing}.
\section{Study objectives}
In this paper, we aim to address the challenges associated with survey-based data collection in information system research. We recognize the potential of Natural Language Processing (NLP) to transform how we gather and analyze data and introduce LLM annotation systems for converting unstructured text data, specifically customer reviews, into analytically useful numeric format. Our research is driven by the goal of exploring the potential of LLM annotation systems as a consistent and scalable means of capturing user attitudes towards technology.

Our research objectives are structured as follows:
\begin{enumerate}
    \item \textbf{Designing an information system:} The initial phase involves designing an LLM annotation system that converts unstructured user reviews into structured annotations reflecting user attitudes based on established technology acceptance models.
    \item \textbf{Consistency validation:} This involves testing whether the LLM annotation system produces consistent results across multiple runs and across different technology acceptance variables. We aim to demonstrate that our LLM-based system can serve as a stable and dependable tool for data annotation.
    \item \textbf{Accuracy validation against human experts:} The final step focuses on evaluating the accuracy of the LLM annotations in comparison to evaluations provided by human experts. Our hypothesis is that expert annotations will concur with those generated by the LLM system.
\end{enumerate}

Through these objectives, our study aims to show that LLM annotation systems can replace traditional questionnaire-based surveys and expert evaluations in technology and information science research. By using ecologically valid data, we want to demonstrate the opportunity for more scalable methods of inferring user attitudes towards technology.
\section{Methods}
\subsection{Designing an Information System for annotating customer reviews with LLMs}
We designed a simple LLM annotation system to evaluate customer reviews using Large Language Models (LLMs) and report their attitudes towards a product in a standardized Technology Acceptance Model (TAM) questionnaire format. Our system operates as follows: Initially, we input a custom prompt that instructs LLMs on annotating reviews based on a specified acceptance model. This prompt also details the required attitude scale (e.g. Likert scale) and the desired output format. In our case, the prompt given to the LLM was as follows:

\begin{quote}
\textit{Your task is to evaluate customer reviews of products based on specific variables that
influence technology acceptance.}

\textit{You will use a 5-point Likert scale to rate each variable in the review.}

\textit{Variables:}
\begin{itemize}
  \item \textit{Performance expectancy,}
  \item \textit{Effort expectancy,}
  \item \textit{Social influence,}
  \item \textit{Facilitating conditions}
\end{itemize}

\textit{Likert Scale:}
\begin{itemize}
  \item \textit{1: The review clearly expresses a negative perception of the factor.}
  \item \textit{2: The review suggests some negative aspects regarding the factor.}
  \item \textit{3: The review does not lean clearly towards a positive or negative perception.}
  \item \textit{4: The review suggests a positive perception of the factor.}
  \item \textit{5: The review clearly expresses a positive perception of the factor.}
\end{itemize}

\textit{If the review does not provide enough information to assess this factor, or the factor is
irrelevant to the context of the review, will assign 0 to the respective variable.}
\end{quote}

This prompt is adaptable to meet researchers' needs. For instance, different variables can be introduced or even defined by the researcher within the prompt. A varied scale (e.g., a 7-point attitude scale) can be utilized, or different category ratings can be included or excluded (e.g., removing the "no information" rating). However, prompt designers must be cautious because the output from the LLM will depend on the prompt given. Although previous research has demonstrated high consistency of results regardless of the prompt wording when it comes to annotations, we maintain that the quality of the prompt is crucial in obtaining the highest quality results \cite{kangale2016mining}.

Upon specifying the prompt, the system is initiated by receiving a list of customer reviews. For this study, we chose 15 refurbished iPhone 13 reviews from the Amazon website, selecting them for their relevance to technology and the extensive user feedback available on the website. Each review is individually processed to extract annotations related to the TAM variables specified in the prompt. This extraction process involves feeding the prompt into the model and parsing the model's response. The parsed responses are standardized into the desired data format. We utilize regular expression syntax to extract this information from the LLM's output and organize it into a data frame. The final result is a well-organized data frame with TAM variables as columns, reviews as rows, and the LLM annotations as elements.

Using a GPT-4 LLM model to generate annotations for 15 selected Amazon iPhone 13 reviews, we obtained the following results in a single run:
\begin{table}[H]
\centering
\begin{tabular}{*{16}{c}}
\toprule
\textbf{Review:} & \textbf{1} & \textbf{2} & \textbf{3} & \textbf{4} & \textbf{5} & \textbf{6} & \textbf{7} & \textbf{8} & \textbf{9} & \textbf{10} & \textbf{11} & \textbf{12} & \textbf{13} & \textbf{14} & \textbf{15} \\
\midrule
Performance expectancy & 2 & 3 & 1 & 1 & 5 & 1 & 5 & 2 & 4 & 2 & 3 & 2 & 3 & 1 & 1 \\
Effort expectancy & 4 & 4 & 1 & 4 & 5 & 1 & 5 & 3 & 2 & 3 & 2 & 1 & 2 & 3 & 1 \\
Social influence & 0 & 0 & 0 & 0 & 0 & 0 & 0 & 0 & 0 & 0 & 0 & 0 & 0 & 1 & 0 \\
Facilitating conditions & 3 & 2 & 1 & 2 & 5 & 1 & 4 & 1 & 2 & 1 & 1 & 2 & 2 & 1 & 1 \\
\bottomrule
\end{tabular}
\end{table}

The workflow generates evaluations as expected, annotating the Likert scale ratings where it thinks the attitude can be annotated and assigning 0 (no information category) where the model thinks that the review does not provide enough information to infer the user's attitude. This is most prominent in the Social Influence variables, where the LLM did not provide attitude ratings for all but one review (review 14).

\subsection{Consistency measure}
We employed the weighted percentage agreement (WPA) to measure the level of consistency between different sets of annotations. This included comparisons between two LLMs annotations, the same LLM across different iterations (Study 1) and LLM and expert annotations (Study 2). WPA calculates the proportion of instances where the annotations matched (agreement), while also taking into account the severity of disagreement. Specifically, smaller disagreements are penalized less than larger disagreements. WPA can take values from 0 to 1 with closer to 1 meaning higher agreement.

The equation for weighted percentage agreement is as follows:
\begin{equation}
WPA = 1 - \frac{\sum_{i = 1}^{n}{w(r_{1i},r_{2i})}}{n \bullet w_{\max}}
\end{equation}
Where:
\begin{itemize}
    \item $n$ is the total number of items (reviews) being compared.
    \item $w(r_{1i},r_{2i})$ represents the penalty weight assigned to the disagreement between the two sets of annotations on the $i$-th item.
    \item $w_{\max}$ is the maximum weight, representing the highest possible penalty for disagreement within the custom weights matrix.
\end{itemize}
The sum $\sum_{i = 1}^{n}{w(r_{1i},r_{2i})}$ calculates the total weighted disagreement over all items and the denominator $n \bullet w_{\max}$ is the maximum weighted disagreement.

The WPA metric requires a weight matrix that contains all possible penalties for the severity of disagreement. We opted for the following penalty matrix:
\[
W = \begin{bmatrix}
0 & 16 & 9 & 4 & 9 & 16 \\
16 & 0 & 1 & 4 & 9 & 16 \\
9 & 1 & 0 & 1 & 4 & 9 \\
4 & 4 & 1 & 0 & 1 & 4 \\
9 & 9 & 4 & 1 & 0 & 1 \\
16 & 16 & 9 & 4 & 1 & 0
\end{bmatrix}
\]

In this matrix, a penalty of 0 denotes complete agreement. Penalties increase in quadratic difference from 1 to 16 for disagreements among annotations 1 through 5. For example, a complete disagreement---annotating a review as clearly negative (1) versus clearly positive (5)---incurs the highest penalty of 16, while disagreements between adjacent annotations (e.g., 1 and 2) receive a minimal penalty of 1.

Annotation of 0 is uniquely treated as a "no information" category and is assigned specific penalties: [0, 16, 9, 4, 9, 16]. Disagreements between a 0 annotation and the extremes (ratings 1 and 5) are penalized most heavily (16), indicating a substantial gap in evaluation between annotating no information about the factor in a review and annotating a clear positive or negative review. Conversely, a comparison of 0 with a neutral annotation (3) attracts the lowest penalty (4), suggesting a somewhat similar lack of decisive positioning, although "neither positive nor negative" is still not the same as "no information about the factor". The penalties for comparisons between 0 and ratings 2 and 4 (9) are moderate, acknowledging that while there is a discernible gap in evaluations, it is less severe than the extreme annotations of 1 or 5.

We determined that other popular metrics of agreement such as Cohen's Kappa, Fleiss' Kappa, and the Intraclass Correlation Coefficient (ICC) were not appropriate for our dataset. Our reason stems from the composition of our data, which blends categorical and ordinal ratings (categorical "No information" rating annotated as 0, and the remaining ratings following a 1 to 5 Likert scale format). Additionally, we observed that some LLM generations (and experts) provided annotations with limited variability and that this could affect the results of more complex agreement metrics, as they require heterogeneous distributions.

\section{Study 1: Consistency validation}
In the first study, we evaluate run-to-run consistency of our workflow, which refers to whether each execution of the described information system produces consistent annotations for the same reviews. Should independent runs yield different annotations, the system's utility would be undermined. It would suggest that the LLM lacks a genuine understanding of the attitudes in the reviews and generates ratings arbitrarily with each execution.

To assess run-to-run consistency, we executed our workflow 50 times on the same set of reviews using two different LLM models (GPT-3.5 and GPT-4). We then calculated the weighted percentage agreement (WPA) for every possible pair of runs (e.g., run 1 with run 2, run 1 with run 3, run 2 with run 3, etc.). The weighted percentage agreement quantifies the proportion of similar annotations across two runs, incorporating a penalty for the severity of disagreement. This process was repeated for each TAM variable. Subsequently, we averaged all possible pairwise agreements to determine an average agreement score for each variable, thereby deriving the consistency ratings across 50 workflow runs.

\subsection{Results}
The assessment of run-to-run consistency yielded the following average weighted percentage agreement (WPA) scores for GPT-3.5: $\widehat{WPA}_{GPT - 3.5\ (PE)} = 0.76$ for Performance Expectancy, $\widehat{WPA}_{GPT - 3.5\ (EE)} = 0.86$ for Effort Expectancy, $\widehat{WPA}_{GPT - 3.5\ (SI)} = 0.68$ for Social Influence, and $\widehat{WPA}_{GPT - 3.5\ (FC)} = 0.66$ for Facilitating Conditions. For GPT-4, the average WPA scores are: $\widehat{WPA}_{GPT - 4\ (PE)} = 0.74$ for Performance Expectancy, $\widehat{WPA}_{GPT - 4\ (EE)} = 0.72$ for Effort Expectancy, $\widehat{WPA}_{GPT - 4\ (SI)} = 0.86$ for Social Influence, and $\widehat{WPA}_{GPT - 4\ (FC)} = 0.76$ for Facilitating Conditions. According to McHugh (2012), these values indicate moderate to strong levels of consistency, where moderate is greater than 0.6 and strong exceeds 0.8 \cite{mchugh2012interrater}. Such results suggest that while the annotations are generally consistent, there are instances where runs produce divergent annotations.

To further analyze these instances of divergence, we compiled  \autoref{tab:table1} and  \autoref{tab:table2}, which present four key statistics for each TAM variable and review: the mode annotation, the range of annotations, the proportion of cases in which the LLM annotated the mode, and the proportion of cases in which the LLM annotated within proximity of the mode (annotation ± 1 from the mode). The methodology for calculating these statistics is detailed in the Supplementary Materials.

\begin{landscape}
\small
\setlength{\tabcolsep}{4pt}

\begin{table}
\caption{Key Statistics for Annotation Consistency across LLM Runs (GPT-3.5)}
\label{tab:table1}
\centering
\begin{tabular}{@{}llcccccccccccccccc@{}}
\toprule
\textbf{Model} & \textbf{Statistic} & \textbf{Review:} & \textbf{1} & \textbf{2} & \textbf{3} & \textbf{4} & \textbf{5} & \textbf{6} & \textbf{7} & \textbf{8} & \textbf{9} & \textbf{10} & \textbf{11} & \textbf{12} & \textbf{13} & \textbf{14} & \textbf{15} \\
\midrule
GPT-3.5 & Mode annotation & PE & 4 & 4 & 1 & 2 & 5 & 1 & 4 & 2 & 4 & 1 & 3 & 2 & 3 & 2 & 1 \\
 & & EE & 3 & 3 & 1 & 2 & 4 & 1 & 4 & 3 & 2 & 1 & 2 & 2 & 3 & 2 & 1 \\
 & & SI & 3 & 0 & 0 & 0 & 4 & 0 & 5 & 0 & 2 & 1 & 2 & 2 & 0 & 1 & 1 \\
 & & FC & 0 & 3 & 1 & 1 & 4 & 1 & 4 & 0 & 2 & 1 & 2 & 2 & 0 & 1 & 1 \\
\cmidrule{2-18}
 & Range & PE & 1 & 0 & 1 & 1 & 0 & 0 & 1 & 2 & 0 & 1 & 2 & 1 & 1 & 1 & 0 \\
 & & EE & 1 & 1 & 1 & 1 & 0 & 0 & 1 & 1 & 0 & 2 & 1 & 1 & 2 & 2 & 1 \\
 & & SI & 3 & 3 & 1 & 4 & 4 & 1 & 5 & 3 & 3 & 3 & 3 & 3 & 3 & 0 & 1 \\
 & & FC & 4 & 4 & 0 & 1 & 1 & 0 & 2 & 4 & 1 & 1 & 1 & 3 & 3 & 2 & 1 \\
\cmidrule{2-18}
 & Proportion of & PE & .58 & 1. & .90 & .74 & 1. & 1. & .70 & .66 & 1. & .86 & .56 & .94 & .90 & .94 & 1. \\
 & mode & EE & .82 & .96 & .90 & .80 & 1. & 1. & .96 & .72 & 1. & .80 & .84 & .76 & .56 & .72 & .96 \\
 & & SI & .56 & .66 & .86 & .54 & .58 & .98 & .56 & .90 & .76 & .94 & .88 & .50 & .88 & 1. & .96 \\
 & & FC & .38 & .42 & 1. & .82 & .82 & 1. & .76 & .40 & .98 & .96 & .80 & .40 & .82 & .60 & .98 \\
\cmidrule{2-18}
 & Proportion in & PE & 1. & 1. & 1. & 1. & 1. & 1. & 1. & .96 & 1. & 1. & 1. & 1. & 1. & 1. & 1. \\
 & proximity of & EE & 1. & 1. & 1. & 1. & 1. & 1. & 1. & 1. & 1. & 1. & 1. & 1. & 1. & 1. & 1. \\
 & mode & SI & .56 & .66 & 1. & .80 & .80 & 1. & .86 & .90 & .76 & .98 & .88 & .56 & .88 & 1. & 1. \\
 & & FC & .40 & .78 & 1. & 1. & 1. & 1. & 1. & .72 & 1. & 1. & 1. & .94 & .90 & 1. & 1. \\
\bottomrule
\end{tabular}
\end{table}

\begin{table}
\caption{Key Statistics for Annotation Consistency across LLM Runs (GPT-4)}
\label{tab:table2}
\centering
\begin{tabular}{@{}llcccccccccccccccc@{}}
\toprule
\textbf{Model} & \textbf{Statistic} & \textbf{Review:} & \textbf{1} & \textbf{2} & \textbf{3} & \textbf{4} & \textbf{5} & \textbf{6} & \textbf{7} & \textbf{8} & \textbf{9} & \textbf{10} & \textbf{11} & \textbf{12} & \textbf{13} & \textbf{14} & \textbf{15} \\
\midrule
GPT-4 & Mode annotation & PE & 3 & 4 & 1 & 2 & 5 & 1 & 5 & 2 & 4 & 2 & 2 & 2 & 3 & 1 & 1 \\
 & & EE & 4 & 3 & 1 & 3 & 5 & 1 & 5 & 3 & 2 & 3 & 4 & 1 & 2 & 2 & 1 \\
 & & SI & 0 & 0 & 0 & 0 & 0 & 0 & 0 & 0 & 0 & 0 & 0 & 0 & 0 & 1 & 0 \\
 & & FC & 4 & 4 & 1 & 1 & 5 & 1 & 5 & 1 & 2 & 1 & 1 & 1 & 2 & 1 & 1 \\
\cmidrule{2-18}
 & Range & PE & 2 & 1 & 0 & 1 & 0 & 0 & 0 & 3 & 1 & 1 & 2 & 1 & 2 & 1 & 0 \\
 & & EE & 4 & 2 & 0 & 3 & 1 & 1 & 1 & 4 & 0 & 3 & 3 & 1 & 4 & 2 & 0 \\
 & & SI & 0 & 0 & 0 & 0 & 0 & 1 & 0 & 0 & 0 & 0 & 0 & 3 & 0 & 1 & 1 \\
 & & FC & 3 & 2 & 1 & 1 & 1 & 0 & 1 & 2 & 2 & 1 & 1 & 2 & 2 & 1 & 0 \\
\cmidrule{2-18}
 & Proportion of & PE & .50 & .74 & 1. & .90 & 1. & 1. & 1. & .90 & .62 & .82 & .92 & .98 & .78 & .88 & 1. \\
 & mode & EE & .54 & .56 & 1. & .48 & .98 & .96 & .70 & .56 & 1. & .48 & .52 & .74 & .34 & .48 & 1. \\
 & & SI & 1. & 1. & 1. & 1. & 1. & .98 & 1. & 1. & 1. & 1. & 1. & .98 & 1. & .96 & .98 \\
 & & FC & .40 & .46 & .98 & .80 & .66 & 1. & .68 & .64 & .88 & .84 & .76 & .46 & .84 & .88 & 1. \\
\cmidrule{2-18}
 & Proportion in & PE & 1. & 1. & 1. & 1. & 1. & 1. & 1. & .98 & 1. & 1. & .98 & 1. & 1. & 1. & 1. \\
 & proximity of & EE & .80 & .86 & 1. & .80 & 1. & 1. & 1. & .90 & 1. & .48 & .84 & 1. & .70 & 1. & 1. \\
 & mode & SI & 1. & 1. & 1. & 1. & 1. & 1. & 1. & 1. & 1. & 1. & 1. & .98 & 1. & 1. & 1. \\
 & & FC & .68 & .78 & 1. & 1. & 1. & 1. & 1. & .92 & 1. & 1. & 1. & .84 & 1. & 1. & 1. \\
\bottomrule
\end{tabular}
\end{table}

\end{landscape}

Key statistics from \autoref{tab:table1} and \autoref{tab:table1} that help identify potential divergences include the range and the proportion of annotations close to the mode. Ideally, annotations should not have a range exceeding 2. A range of 2 or less suggests a desired, narrow clustering of annotations. However, the range metric alone does not account for the frequency of deviations; a single outlier can easily push the range above 2. The proportion of annotations in proximity of the mode considers the percentage of annotations within a 2-unit range of the mode, providing a more nuanced view. Even in cases where a single outlier expands the range above 2, if the majority of annotations cluster around the mode (within this 2-unit proximity), then this metric will remain high. We argue that values above 0.9 reflect consistent annotations, whereas values below 0.8 suggest inconsistency.

Our results indicate that achieving perfect consistency in LLM annotations across multiple runs, variables, and stimuli (reviews) is challenging. No review (stimulus) has shown perfect consistency across runs. Only review 15 came closest to achieving near-perfect consistency in annotations, both for GPT-3.5 and GPT-4.

In a comparative analysis, the GPT-4 model shows a slightly more consistent performance than GPT-3.5, indicated by less variability in the range and a higher proportion of annotations near the mode. However, this advantage is small and not consistent across all variables. GPT-4 performs better in annotating Social Influence but slightly worse in Effort Expectancy compared to GPT-3.5. Analyzing the weighted proportion agreement (WPA) between the mode annotations of GPT-3.5 and GPT-4 for all variables, we find: $\widehat{WPA}_{Between\ LLMs\ (PE)} = 0.98$ for Performance Expectancy, $\widehat{WPA}_{Between\ LLMs\ (EE)} = 0.94$ for Effort Expectancy, $\widehat{WPA}_{Between\ LLMs\ (SI)} = 0.50$ for Social Influence and $\widehat{WPA}_{Between\ LLMs\ (PE)} = 0.81$ for Facilitating Conditions. These findings indicate that for core technology acceptance variables like Performance Expectancy and Effort Expectancy, GPT-3.5 and GPT-4 produce remarkably similar annotations (the mode). The annotations for Facilitating Conditions also show considerable consistency. In contrast, annotations for Social Influence reveal some inconsistency between the models.

\subsection{Improving consistency with model temperature}
Previous research on LLM annotation systems has shown that the consistency of these models can be improved by adjusting the model temperature \cite{reiss2023testing}. Model temperature refers to a hyperparameter that controls the diversity of the output generated by the model. A standard setting for many LLM models is a temperature of 1. This is considered the "default" or "neutral" setting, where the model generates text that balances randomness and predictability. Lowering the model temperature increases the model's confidence in its predictions, leading to outputs that are less random and more predictable. However, setting the temperature too low can result in the model producing text that is repetitive or overly cautious. While a very low temperature is generally avoided in text generation due to its limiting effect on model performance, it could be beneficial for annotation systems where consistency is key. Although a temperature of 0 may be too low, potentially compromising the quality of model annotations due to a lack of randomness. Randomness in LLM outputs, introduced by higher temperature settings, allows the model to explore different possibilities and generate more diverse responses. This diversity can lead to improved reasoning and output quality, as the model considers multiple perspectives and avoids being overly deterministic. However, in the case of annotation systems, where consistency is a primary goal, reducing randomness by lowering the temperature can be advantageous. In this paper, we explore a temperature setting of 0.25 to assess its impact on improving annotation consistency and to see whether it influences the mode of annotations. \autoref{tab:table3} displays the outcomes for an annotation system using the GPT-3.5 model at a reduced temperature setting.

\begin{landscape}
\small
\setlength{\tabcolsep}{4pt}

\begin{table}
\caption{Key Statistics for Annotation Consistency with Reduced Temperature (GPT-3.5, Temperature 0.25)}
\label{tab:table3}
\centering
\begin{tabular}{@{}lcccccccccccccccc@{}}
\toprule
\textbf{Statistic} & \textbf{Review:} & \textbf{1} & \textbf{2} & \textbf{3} & \textbf{4} & \textbf{5} & \textbf{6} & \textbf{7} & \textbf{8} & \textbf{9} & \textbf{10} & \textbf{11} & \textbf{12} & \textbf{13} & \textbf{14} & \textbf{15} \\
\midrule
Mode annotation & PE & 4 & 4 & 1 & 2 & 5 & 1 & 4 & 2 & 4 & 1 & 3 & 2 & 3 & 2 & 1 \\
 & EE & 3 & 3 & 1 & 2 & 4 & 1 & 4 & 3 & 2 & 1 & 2 & 2 & 3 & 2 & 1 \\
 & SI & 3 & 0 & 0 & 0 & 4 & 0 & 5 & 0 & 3 & 0 & 0 & 3 & 0 & 1 & 0 \\
 & FC & 3 & 3 & 1 & 1 & 4 & 1 & 4 & 4 & 2 & 1 & 2 & 2 & 0 & 0 & 1 \\
\midrule
Range & PE & 1 & 0 & 0 & 1 & 0 & 0 & 1 & 0 & 0 & 1 & 2 & 1 & 1 & 0 & 0 \\
 & EE & 1 & 0 & 0 & 0 & 0 & 0 & 0 & 0 & 0 & 2 & 1 & 1 & 1 & 1 & 0 \\
 & SI & 3 & 3 & 1 & 3 & 4 & 0 & 1 & 0 & 3 & 0 & 0 & 3 & 0 & 0 & 0 \\
 & FC & 3 & 4 & 0 & 1 & 1 & 0 & 1 & 4 & 0 & 1 & 1 & 3 & 2 & 2 & 0 \\
\midrule
Proportion of & PE & .88 & 1. & 1. & .98 & 1. & 1. & .80 & 1. & 1. & .94 & .88 & .98 & .98 & 1. & 1. \\
mode & EE & .90 & 1. & 1. & 1. & 1. & 1. & 1. & 1. & 1. & .90 & .98 & .86 & .68 & .98 & 1. \\
 & SI & .76 & .70 & .98 & .44 & .50 & 1. & .82 & 1. & .88 & 1. & 1. & .64 & 1. & 1. & 1. \\
 & FC & .72 & .62 & 1. & .74 & .86 & 1. & .98 & .40 & 1. & .98 & .92 & .58 & .98 & .50 & 1. \\
\midrule
Frequency in & PE & 1. & 1. & 1. & 1. & 1. & 1. & 1. & 1. & 1. & 1. & 1. & 1. & 1. & 1. & 1. \\
proximity of & EE & 1. & 1. & 1. & 1. & 1. & 1. & 1. & 1. & 1. & 1. & 1. & 1. & 1. & 1. & 1. \\
mode & SI & .76 & .70 & 1. & .80 & .84 & 1. & 1. & 1. & .88 & 1. & 1. & .64 & 1. & 1. & 1. \\
 & FC & .76 & .86 & 1. & 1. & 1. & 1. & 1. & .40 & 1. & 1. & 1. & .98 & .98 & .98 & 1. \\
\bottomrule
\end{tabular}
\end{table}

\end{landscape}

The results show that decreasing the model's temperature leads to improved overall consistency in annotations. Variables such as Performance Expectancy and Effort Expectancy were annotated with almost perfect consistency, while the annotations for two other variables also saw significant improvements. Nevertheless, we observed minor changes in the mode of annotations. For instance, with a temperature of 1, the GPT-3.5 model categorized review 8 under the variable Facilitating Conditions as 0 (indicating no information), whereas at a temperature of 0.25, the same model assessed it as 4 (reflecting a strong positive attitude towards Facilitating conditions). This outcome becomes less surprising upon examining the range and frequencies of ratings across both runs, which reveal that this particular review is quite difficult to annotate, producing a diverse range of ratings in both instances.

\section{Study 2: Accuracy validation against human experts}
In the second study, we compared annotations derived from LLMs versus from human experts. We enlisted three independent experts who hold PhD degrees in information systems research and have published works on Technology Acceptance Models (the experts were not involved in the authorship of this paper). We presented them with the prompt from part one and asked them to evaluate the same 15 Amazon iPhone 13 reviews. Each review was annotated based on four Unified Theory of Acceptance and Use of Technology (UTAUT) variables, using a Likert scale. We then compared their annotations to those generated by LLMs.

\subsection{Results}
\autoref{tab:table4} presents the agreement between the evaluations of three human experts and mode evaluations generated by the GPT-4 model (mode evaluations taken from Table 2). The average WPA across all pairs (experts and LLM) is highest for Performance Expectancy ($\widehat{WPA}_{PE} = 0.87$) and Effort Expectancy ($\widehat{WPA}_{EE} = 0.73$). The mean WPA scores for Social Influence and Facilitating Conditions are on the threshold of moderate and weak, revealing a significantly lower level of consensus when it comes to annotating these factors in reviews ($\widehat{WPA}_{SI} = 0.67$ and $\widehat{WPA}_{FC} = 0.60$).

\begin{table}[H]
\centering
\caption{Agreement between Human Experts and GPT-4 Annotations}
\label{tab:table4}
\begin{tabular}{lc}
\toprule
\textbf{Variable / Comparison} & \textbf{Agreement Score} \\
\midrule
\textbf{Performance Expectancy (PE)} & \\
Expert 1 vs Expert 2 & .6625 \\
Expert 1 vs Expert 3 & .9542 \\
Expert 1 vs GPT-4 & \textbf{.9458} \\
Expert 2 vs Expert 3 & .6833 \\
Expert 2 vs GPT-4 & \textbf{.6917} \\
Expert 3 vs GPT-4 & \textbf{.9833} \\
\textbf{Effort Expectancy (EE)} & \\
Expert 1 vs Expert 2 & .6625 \\
Expert 1 vs Expert 3 & .8250 \\
Expert 1 vs GPT-4 & \textbf{.6667} \\
Expert 2 vs Expert 3 & .6625 \\
Expert 2 vs GPT-4 & \textbf{.6542} \\
Expert 3 vs GPT-4 & \textbf{.8750} \\
\textbf{Social Influence (SI)} & \\
Expert 1 vs Expert 2 & .2917 \\
Expert 1 vs Expert 3 & .8208 \\
Expert 1 vs GPT-4 & \textbf{.9333} \\
Expert 2 vs Expert 3 & .4208 \\
Expert 2 vs GPT-4 & \textbf{.3250} \\
Expert 3 vs GPT-4 & \textbf{.7542} \\
\textbf{Facilitating Conditions (FC)} & \\
Expert 1 vs Expert 2 & .4583 \\
Expert 1 vs Expert 3 & .4750 \\
Expert 1 vs GPT-4 & \textbf{.3083} \\
Expert 2 vs Expert 3 & .6750 \\
Expert 2 vs GPT-4 & \textbf{.6500} \\
Expert 3 vs GPT-4 & \textbf{.8417} \\
\bottomrule
\end{tabular}
\end{table}

We can categorize the average WPA for each variable into two groups: the mean agreement among human experts alone (comparing expert 1 with expert 2, 2 with 3, and 1 with 3) and the mean agreement between all human experts and the LLM model (evaluations of experts 1, 2, and 3 with GPT-4). This allows us to assess if there is a significant difference between the consensus among human experts and their agreement with LLMs evaluations. For Performance Expectancy, the mean agreement between human experts was $\widehat{WPA}_{Between\ Human\ Experts\ (PE)} = 0.77$ while their agreement with GPT-4 evaluations was slightly higher at $\widehat{WPA}_{Human\ Experts\ with\ GPT - 4\ (PE)} = 0.87$. This minor difference suggests that, on average, the evaluations provided by the LLM closely match those of the human experts. A similar pattern is observed for Effort Expectancy, where the agreement between human experts and between humans and the AI does not significantly diverge ($\widehat{WPA}_{Between\ Human\ Experts\ (EE)} = 0.72$, $\widehat{WPA}_{Human\ Experts\ with\ GPT - 4\ (EE)} = 0.73$). The trend holds for Facilitating Conditions as well, with agreements of ($\widehat{WPA}_{Between\ Human\ Experts\ (FC)} = 0.54$, $\widehat{WPA}_{Human\ Experts\ with\ GPT - 4\ (FC)} = 0.60$). However, a notable exception is found in the case of Social Influence, where the agreement between human experts and the AI ($\widehat{WPA}_{Human\ Experts\ with\ GPT - 4\ (SI)} = 0.67$) significantly surpasses the agreement among human experts alone ($\widehat{WPA}_{Between\ Human\ Experts\ (SI)} = 0.51$). This implies that, on average, human experts align more closely with the evaluations provided by LLMs than they do with each other's assessments.

\section{Discussion}
Technology acceptance is a useful measure that helps companies design and target their products effectively. While consumer data has become increasingly accessible, the opportunity for analysis of consumer trends was previously limited by the reliance on traditional surveys and human expert reviewers. The development of Natural Language Processing with Large Language Model (LLM) annotation systems has provided researchers and companies with a powerful new tool for measuring technology acceptance in a scalable and cost-effective manner.

In this paper, we demonstrated the success of LLMs, specifically GPT models, in this field. We find that LLM's reliability and agreement with human experts are largely sufficient for current analysis tasks. Furthermore, as shown by the agreement scores, LLMs can be as useful as human experts when it comes to annotating attitudes in customer data. In our study, LLMs performed exceptionally well, with agreement scores between the AI and human experts coming very close to or even surpassing the agreement between human experts themselves. For example, we found that for Performance Expectancy, Facilitating Conditions, and Social Influence, the agreement is higher between the experts and the LLM than between the experts themselves. For the remaining category, Effort Expectancy, the agreement scores are almost identical. Statistically, a higher agreement score of the LLMs and experts means that scores generated by the LLM are more closely clustered around the mean of the human reviewers than the human reviews themselves. Such a result suggests that LLMs can be reliably and accurately used to analyze and annotate consumer data.

The use of LLM annotation systems offers several advantages over traditional survey methods, including cost-effectiveness, scalability, and the ability to leverage existing user-generated data from various online sources. Of these advantages, the most significant is the nearly unlimited LLM capacity to analyze reviews. For example, our analysis was limited by the time constraints of human reviewers with 15 reviews an acceptable middle ground between statistical power and feasibility per human reviewers. While the comparatively lower number of reviews does limit our analysis, it also shows the increased/superior capabilities of LLM analysis compared to traditional methods. However, it is important to note several potential limitations and challenges associated with LLM annotation systems, such as the need for prompt engineering, the closed-source nature of many popular LLMs, and the trade-off between consistency and flexibility.

Our findings have significant implications for researchers and practitioners in the field of information systems and technology acceptance. LLM annotation systems could be integrated into research and product development processes, providing a valuable tool for understanding user attitudes and preferences, refining technology design, and conducting market research. Despite these promising results, further research and development in this area are needed. This includes exploring different LLM models, fine-tuning techniques, and evaluating the performance of LLM annotation systems across different domains and technology products. Additionally, ethical considerations and potential risks associated with the use of LLM annotation systems, such as ethical data scraping, ethical model selection, and the need for transparency and accountability of LLMs, should be carefully addressed. With improved tuning, such as modifying the temperature or training models precisely for annotation tasks, LLMs are likely to be widely used to analyze the significant amounts of online consumer data. Our study shows that such an approach will likely be successful and should be an area of interest and further research, enabling deeper insights into technology design and adoption.

\section{Recommendations for LLM Annotation Systems Use}
Our research has demonstrated that Large Language Model (LLM) annotation systems can be a reliable and consistent tool for inferring user attitudes from unstructured text data. However, we observed some instances of small divergences from the mode in single-run annotations. To address this issue and ensure the highest quality of annotations, we propose the following recommendations for researchers interested in implementing LLM annotation systems in their work.

We advise researchers to consider the mode of LLM annotations across multiple runs as the true sentiment inferred by the LLM model. While single-run annotations may exhibit minor variations, the mode represents the most frequently occurring annotation and is likely to be the most accurate representation of the model's understanding of the text. To obtain the mode annotation, we recommend executing at least 10, preferably more, runs of the same LLM algorithm on the data. By conducting multiple runs, researchers can mitigate the impact of randomness imposed by the model temperature and obtain a more stable and reliable annotation. As demonstrated in our study, reducing the temperature to lower values, such as 0.25, can improve consistency without diminishing the model's reasoning capabilities. For example, when using the GPT-3.5 model with a temperature of 0.25, we observed that the proportion of annotations matching the mode increased to 100\% for Performance Expectancy and Effort Expectancy across all reviews.

In addition to the mode, other metrics such as annotation range and proportions of the mode or proximity to the mode can be used to investigate the correctness and consistency of annotations. We recommend that future researchers employ these metrics in their implementations of LLM annotation systems. By examining the range of annotations, researchers can identify instances where the LLM model produces divergent annotations, potentially indicating areas of uncertainty or ambiguity in the text. Similarly, calculating the proportion of annotations that match or are close to the mode can provide a measure of the system's consistency and reliability.

In conclusion, we strongly recommend that future researchers employing LLM annotation systems in their work adopt a multi-run approach, using the mode annotation as the true sentiment inferred by the model. By combining this approach with an examination of additional metrics and appropriate temperature settings, researchers can ensure the highest quality and consistency of annotations, unlocking the full potential of LLM annotation systems in technology acceptance research and beyond.

\textbf{Supplementary Materials:} Additional details on the methodology and full numerical results are available in the supplementary materials at \url{https://osf.io/pk63g/}.

\end{document}